\documentclass{article}
\usepackage[utf8]{inputenc}
\usepackage[margin=20truemm]{geometry}
\usepackage{amsmath}
\usepackage{amssymb}
\usepackage{mathtools}
\usepackage{amsthm}
\usepackage[colorlinks=true,linkcolor=black,anchorcolor=black,citecolor=black,filecolor=black,menucolor=black,runcolor=black,urlcolor=black]{hyperref}
\usepackage{booktabs}
\usepackage{multirow}
\usepackage{natbib}

\title{\textbf{Subsampling for Knowledge Graph Embedding Explained}}
\author{\textbf{Hidetaka Kamigaito}\footnote{Nara Institute of Science and Technology (NAIST), Nara, Japan. kamigaito.h@is.naist.jp}\hspace{1mm} and \textbf{Katsuhiko Hayashi}\footnote{Hokkaido University, Hokkaido, Japan.}}
\date{}

\begin{document}

\maketitle

\begin{abstract}
    In this article, we explain the recent advance of subsampling methods in knowledge graph embedding (KGE) starting from the original one used in word2vec.
\end{abstract}

\section{Negative Sampling Loss}
Knowledge graph completion (KGC) is a research topic for automatically inferring new links in a KG that are likely but not yet known to be true.

We denote a triplet representing entities $e_{i}$, $e_{j}$ and their relation $r_{k}$ as $(e_{i},r_{k},e_{j})$. In a typical KGC task, the model receives a query $(e_{i},r_{k},?)$ or $(?,r_{k},e_{j})$ and predicts the entity corresponding to $?$.

Knowledge graph embedding (KGE) is a well-known scalable approach for KGC.
In KGE, a KGE model scores a triplet $(e_{i},r_{k},e_{j})$ by using a scoring function $s_{\theta}(x,y)$.
Due to the computational cost, training of $s_{\theta}(x,y)$ commonly relies on the following negative sampling loss function \citep{rotate,ahrabian-etal-2020-structure}:
\begin{equation}
 \ell_{base}=-\frac{1}{|D|}\sum_{(x,y) \in D} \Bigl[\log(\sigma(s_{\theta}(x,y)+\gamma)) + \frac{1}{\nu}\sum_{y_{i}\sim p_n(y_{i}|x)}^{\nu}\log(\sigma(-s_{\theta}(x,y_i)-\gamma))\Bigr],\label{eq:ns:loss:kge}
\end{equation}
where $D=\{(x_{1},y_{1}),\cdots, (x_{n},y_{n})\}$ represents observables that follow $p_{d}(x,y)$, $p_n(y|x)$ is the noise distribution, $\sigma$ is the sigmoid function, $\nu$ is the number of negative samples per positive sample $(x,y)$, and $\gamma$ is a margin term.

\section{Subsampling in Negative Sampling Loss}

Eq. (\ref{eq:ns:loss:kge}) is on the assumption that the NS loss function fits the model to the distribution $p_{d}(y|x)$ defined from the observed data. However, what the NS loss actually does is to fit the model to the true distribution $p'_{d}(y|x)$ that exists behind the observed data.
To fill in the gap between $p_{d}(y|x)$ and $p'_{d}(y|x)$, \cite{icml2022erratum,pmlr-v162-kamigaito22a} theoretically add $A(x,y)$ and $B(x)$ to Eq. (\ref{eq:ns:loss:kge}) as follows\footnote{We include the detailed derivation of this function in Appendix \ref{app:derivation}.}:
\begin{equation}
    \ell_{sub}=-\frac{1}{|D|}\sum_{(x,y) \in D} \Bigl[A(x,y)\log(\sigma(s_{\theta}(x,y)+\gamma)) +\frac{1}{\nu}\sum_{y_{i}\sim p_n(y_{i}|x)}^{\nu}B(x)\log(\sigma(-s_{\theta}(x,y_i)-\gamma))\Bigr].
\label{eq:subsamp}
\end{equation}
In this formulation, we can consider several assumptions for deciding $A(x,y)$ and $B(x)$. We introduce the assumptions in the following subsections. 

\subsection{Subsampling in word2vec (Base)}

As a basic subsampling approach, \cite{rotate} used the original word2vec-based method for KGE learning defined as follows:
\begin{equation}
 A(x,y)=B(x,y)=\frac{\frac{1}{\sqrt{\#(x,y)}}|D|}{\sum_{(x',y') \in D}\frac{1}{\sqrt{\#(x',y')}}},
 \label{eq:subsamp:default}
\end{equation}
where $\#$ is the symbol for frequency and $\#(x,y)$ represents the frequency of $(x,y)$\footnote{In the original word2vec, they randomly discard a word by a probability $1-\sqrt{\frac{t}{f}}$, where $t$ is a constant value and $f$ is a frequencty of a word. This is similar to randomly keep a word with a probability $\sqrt{\frac{t}{f}}$.}.
Note that the actual $(x,y)$ occurs at most once in the KG, so when $(x,y)=(e_{i},r_{k},e_{j})$, they approximate the frequency of $(x,y)$ as follows:
\begin{equation}
 \#(x,y) \approx \#(e_{i},r_{k})+\#(r_{k},e_{j}).
 \label{eq:subsamp:approx}
\end{equation}

Different from the form in Eq. (\ref{eq:subsamp}), Eq. (\ref{eq:subsamp:default}) use $A(x,y)$ and $B(x,y)$, instead of $A(x,y)$ and $B(x)$. Thus, their approach does not follow the theoretically induced loss function in Eq. (\ref{eq:subsamp}).

\subsection{Frequency-based Subsampling (Freq)}

Frequency-based subsampling \citep{pmlr-v162-kamigaito22a} is based on the assumption that in $p'_{d}(y|x)$, $(x,y)$ originally has a frequency, but the observed one is at most 1.
Since $A(x,y)$ needs to discount the frequency of $(x,y)$, and $B(x)$ needs to discount that of $x$, we can derive the following subsampling method based on word2vec \citep{ns} as implemented by the previous work \citep{rotate}\footnote{\url{https://github.com/DeepGraphLearning/KnowledgeGraphEmbedding}}:

\begin{equation}
 A(x,y)=\frac{\frac{1}{\sqrt{\#(x,y)}}|D|}{\sum_{(x',y') \in D}\frac{1}{\sqrt{\#(x',y')}}}, B(x)=\frac{\frac{1}{\sqrt{\#x}}|D|}{\sum_{x' \in D}\frac{1}{\sqrt{\#x'}}}.
 \label{eq:subsamp:freq}
\end{equation}

\begin{table}[t!]
\caption{Evaluation results of \cite{pmlr-v162-kamigaito22a} for each subsampling method on the FB15k-237, WN18RR, and YAGO3-10 datasets. \textit{Sub.} denotes subsampling, \textit{None} denotes model that did not use subsampling, \textit{Base} denotes Eq. (\ref{eq:subsamp:default}), \textit{Freq} denotes Eq. (\ref{eq:subsamp:freq}), and \textit{Uniq} denotes Eq. (\ref{eq:subsamp:uniq}).\label{tab:sub}}
\vspace{2.5mm}
\centering
\small
\begin{tabular}{llcccccccccccc}
\toprule
\multirow{3}{*}{\textbf{Model}}&\multirow{3}{*}{ \textbf{Sub}.}&\multicolumn{4}{c}{\textbf{FB15k-237}} & \multicolumn{4}{c}{\textbf{WN18RR}} & \multicolumn{4}{c}{\textbf{YAGO3-10}}                   \\
\cmidrule(lr){3-6}\cmidrule(lr){7-10}\cmidrule(lr){11-14}
&&\multirow{2}{*}{\textbf{MRR}}&\multicolumn{3}{c}{\textbf{Hits@}}&\multirow{2}{*}{\textbf{MRR}}&\multicolumn{3}{c}{\textbf{Hits@}}&\multirow{2}{*}{\textbf{MRR}}&\multicolumn{3}{c}{\textbf{Hits@}} \\
\cmidrule(lr){4-6}\cmidrule(lr){8-10}\cmidrule(lr){12-14}
                        & &                      &  \textbf{1}     & \textbf{3}     & \textbf{10}&                     &  \textbf{1}     & \textbf{3}     & \textbf{10}&                     &  \textbf{1}     & \textbf{3}     & \textbf{10}    \\
\cmidrule(r){1-2}\cmidrule(lr){3-6}\cmidrule(lr){7-10}\cmidrule(lr){11-14}
\multirow{4}{*}{\textbf{RESCAL}} & None & 17.2 & 9.9 & 18.1 & 31.8 & 41.5 & 39.0 & 42.3 & 45.9 & - & - & - & - \\
 & Base & 22.3 & 13.9 & 24.2 & 39.8 & 43.3 & 40.7 & 44.5 & 48.2 & - & - & - & - \\
 & Freq & \textbf{26.6} & 17.4 & \textbf{29.4} & \textbf{45.1}& \textbf{44.1} & 41.1 & \textbf{45.6} & \textbf{49.5} & - & - & - & - \\
 & Uniq & \textbf{26.6} & \textbf{17.6} & 29.3 & 44.9 & \textbf{44.1} & \textbf{41.4} & 45.5 & \textbf{49.5} & - & - & - & - \\
\cmidrule(r){1-2}\cmidrule(lr){3-6}\cmidrule(lr){7-10}\cmidrule(lr){11-14}
\multirow{4}{*}{\textbf{ComplEx}} & None & 22.4 & 14.0 & 24.2 & 39.5 & 45.0 & 40.9 & 46.6 & 53.4 & - & - & - & -\\
 & Base & 32.2 & 23.0 & 35.1 & 51.0 & 47.1 & 42.8 & 48.9 & 55.7 & - & - & - & - \\
 & Freq & \textbf{32.8} & \textbf{23.6} & \textbf{36.1} & 51.2 & \textbf{47.6} & \textbf{43.3} & 49.3 & \textbf{56.3} & - & - & - & - \\
 & Uniq & 32.7 & 23.5 & 35.8 & \textbf{51.3} & \textbf{47.6} & 43.2 & \textbf{49.5} & \textbf{56.3} & - & - & - & - \\
\cmidrule(r){1-2}\cmidrule(lr){3-6}\cmidrule(lr){7-10}\cmidrule(lr){11-14}
\multirow{4}{*}{\textbf{DistMult}} & None & 22.2 & 14.0 & 24.0 & 39.4 & 42.4 & 38.3 & 43.6 & 51.0 & - & - & - & - \\
 & Base & \textbf{30.8} & \textbf{22.1} & \textbf{33.6} & \textbf{48.4} & 43.9 & 39.4 & 45.2 & 53.3 & - & - & - & - \\
 & Freq & 29.9 & 21.2 & 32.7 & 47.5 & \textbf{44.6} & \textbf{40.0} & 45.9 & \textbf{54.4} & - & - & - & -\\
 & Uniq & 29.1 & 20.3 & 31.8 & 46.6 & \textbf{44.6} & 39.9 & \textbf{46.2} & 54.3 & - & - & - & - \\
\cmidrule(r){1-2}\cmidrule(lr){3-6}\cmidrule(lr){7-10}\cmidrule(lr){11-14}
\multirow{4}{*}{\textbf{TransE}} & None & 33.0 & 22.8 & 37.2 & \textbf{53.0} & 22.6 & 1.8 & 40.1 & 52.3 & 50.6 & 40.9 & 56.6 & 67.7\\
 & Base & 32.9 & 23.0 & 36.8 & 52.7& 22.4 & 1.3 & 40.1 & 53.0 & 51.2 & 41.5 & \textbf{57.6} & \textbf{68.3} \\
 & Freq & \textbf{33.6} & \textbf{24.0} & \textbf{37.3} & 52.9& 23.0 & 1.9 & 40.7 & \textbf{53.7} & 51.3 & 41.9 & 57.2 & 68.1 \\
 & Uniq & 33.5 & 23.9 & \textbf{37.3} & 52.8 & \textbf{23.2} & \textbf{2.2} & \textbf{41.0} & 53.4 & \textbf{51.4} & \textbf{42.0} & \textbf{57.6} & 67.9\\
\cmidrule(r){1-2}\cmidrule(lr){3-6}\cmidrule(lr){7-10}\cmidrule(lr){11-14}
\multirow{4}{*}{\textbf{RotatE}} & None & 33.1 & 23.1 & 37.1 & 53.1 & 47.3 & 42.6 & 49.1 & 56.7 & 50.6 & 41.1 & 56.5 & 67.8\\
 & Base & 33.6 & 23.9 & 37.4 & \textbf{53.2} & 47.6 & 43.1 & 49.5 & 56.6 & 50.8 & 41.8 & 56.5 & 67.6 \\
 & Freq & \textbf{34.0} & \textbf{24.5} & \textbf{37.6} & \textbf{53.2} & 47.8 & 42.9 & \textbf{49.8} & \textbf{57.4} & 51.0 & 41.9 & 56.5 & 67.8 \\
 & Uniq & \textbf{34.0} & \textbf{24.5} & \textbf{37.6} & 53.0 & \textbf{47.9} & \textbf{43.5} & 49.6 & 56.7 & \textbf{51.5} & \textbf{42.5} & \textbf{56.8} & \textbf{68.3}\\
\cmidrule(r){1-2}\cmidrule(lr){3-6}\cmidrule(lr){7-10}\cmidrule(lr){11-14}
\multirow{4}{*}{\textbf{HAKE}} & None & 32.3 & 21.6 & 36.9 & 53.2 & 49.1 & 44.5 & 51.1 & 57.8 & 53.4 & 44.9 & 58.7 & 68.4\\
 & Base & 34.5 & 24.7 & 38.2 & 54.3 & \textbf{49.8} & 45.3 & \textbf{51.6} & 58.2 & 54.3 & 46.1 & 59.5 & 69.2\\
 & Freq & 34.9 & 25.2 & 38.6 & 54.2 & 49.7 & 45.2 & 51.4 & \textbf{58.5} & 54.0 & 45.5 & 59.4 & 69.1\\
 & Uniq & \textbf{35.4} & \textbf{25.8} & \textbf{38.9} & \textbf{54.7} & \textbf{49.8} & \textbf{45.4} & 51.5 & 58.3 & \textbf{55.0} & \textbf{46.6} & \textbf{60.1} & \textbf{69.8}\\
                      \bottomrule
\end{tabular}
\end{table}

\subsection{Unique-based Subsampling (Uniq)}

In the true distribution $p'_{d}(y|x)$, however, if we assume that $(x,y)$ has frequency $1$ at most, as in the observation, then $p'_d(y|x)=p'_d(x,y)/p'_d(x) \propto 1/p'_d(x)$, so $p'_d(y|x)$ is the same for an $x$ independent from $y$. Therefore, under this assumption, we have only need to consider a discount for $p_d(x)$ and can derive the unique-based subsampling \citep{pmlr-v162-kamigaito22a} as follows:
\begin{equation}
 A(x,y)=B(x)=\frac{\frac{1}{\sqrt{\#x}}|D|}{\sum_{x' \in D}\frac{1}{\sqrt{\#x'}}}.
 \label{eq:subsamp:uniq}
\end{equation}

\section{Effectiveness of Subsampling in KGE}

We conducted experiments to evaluate our subsampling methods. We used FB15k-237~\citep{fb15k-237}, WN18RR, and YAGO3-10~\citep{conve} for the evaluation. As comparison methods, we used ComplEx~\citep{complex}, RESCAL~\citep{rescal}, DistMult~\citep{distmult}, TransE~\citep{transe}, RotatE~\citep{rotate}, and HAKE~\citep{hake}. We followed the original settings of~\cite{rotate} for ComplEx, DistMult, TransE, and RotatE with their implementation\footnote{\url{https://github.com/DeepGraphLearning/KnowledgeGraphEmbedding}} and the original settings of~\cite{hake} for HAKE with their implementation\footnote{\url{https://github.com/MIRALab-USTC/KGE-HAKE}}. In RESCAL, we inherited the original setting of DistMult and set the dimension size to 500 for saving computational time.
Since \cite{unified} refer to the smoothing effect of self-adversarial negative sampling (SANS)~\citep{rotate} that is a role of subsampling, we applied subsampling on SANS for investigating the performance in practical settings.

Table \ref{tab:sub} shows the result. We can see that subsampling improved KG completion performances from the methods without subsampling. Furthermore, frequency-based and unique-based subsampling basically outperformed the baseline subsampling.

\bibliographystyle{plainnat}
\bibliography{main}

\clearpage
\appendix

\section{The Detailed Derivation of Eq.~(\ref{eq:subsamp})}
\label{app:derivation}
We can reformulate the NS loss in Eq.~(\ref{eq:ns:loss:kge}) as follows:
\begin{align}
(\ref{eq:ns:loss:kge})=& -\frac{1}{|D|}\sum_{(x,y) \in D} \Bigl[\log(\sigma(s_{\theta}(x,y)+\gamma)) + \frac{1}{\nu}\sum_{y_{i}\sim p_n(y_{i}|x)}^{\nu}\log(\sigma(-s_{\theta}(x,y_i)-\gamma))\Bigr]\nonumber\\
= & -\frac{1}{|D|}\sum_{(x,y) \in D} \log(\sigma(s_{\theta}(x,y)+\gamma)) -\frac{1}{|D|}\sum_{(x,y) \in D}\frac{1}{\nu}\sum_{y_{i}\sim p_n(y_{i}|x)}^{\nu}\log(\sigma(-s_{\theta}(x,y_i)-\gamma)).
\label{eq:app:sub:intro}
\end{align}
Here, we can consider the following approximation based on the Monte Carlo method:
\begin{equation}
    \frac{1}{\nu}\sum_{y_{i}\sim p_n(y_{i}|x)}^{\nu}\log(\sigma(-s_{\theta}(x,y_i)-\gamma)) \approx \sum_{y}p_n(y|x)\log(\sigma(-s_{\theta}(x,y)-\gamma)).
    \label{eq:app:sub:intro:approx}
\end{equation}
Using Eq.~(\ref{eq:app:sub:intro:approx}), we can reformulate Eq.~(\ref{eq:app:sub:intro}) as follows: 
\begin{equation}
(\ref{eq:app:sub:intro}) \approx -\frac{1}{|D|}\sum_{(x,y) \in D} \log(\sigma(s_{\theta}(x,y)+\gamma)) -\frac{1}{|D|}\sum_{(x,y) \in D}\sum_{y'}p_n(y'|x)\log(\sigma(-s_{\theta}(x,y')-\gamma)).
\label{eq:app:sub:intro2}
\end{equation}
Similr to Eq.~(\ref{eq:app:sub:intro:approx}), we can consider the following approximation by the the Monte Carlo method:
\begin{align}
    -&\frac{1}{|D|}\sum_{(x,y) \in D} \log(\sigma(s_{\theta}(x,y)+\gamma))\approx -\sum_{x,y}\log(\sigma(s_{\theta}(x,y)+\gamma))p_d(x,y), \nonumber\\
    -&\frac{1}{|D|}\sum_{(x,y) \in D}\sum_{y'}p_n(y'|x)\log(\sigma(-s_{\theta}(x,y')-\gamma))\approx -\sum_{x}\sum_{y'}p_n(y'|x)\log(\sigma(-s_{\theta}(x,y')-\gamma))p_d(x).
    \label{eq:app:sub:approx2}
\end{align}
Using Eq.~(\ref{eq:app:sub:approx2}), we can reformulate Eq.~(\ref{eq:app:sub:intro2}) as follows:
\begin{align}
(\ref{eq:app:sub:intro2}) \approx & -\sum_{x,y}\log(\sigma(s_{\theta}(x,y)+\gamma))p_d(x,y) -\sum_{x}\sum_{y'}p_n(y'|x)\log(\sigma(-s_{\theta}(x,y')-\gamma))p_d(x)\nonumber\\
= & -\sum_{x,y}\log(\sigma(s_{\theta}(x,y)+\gamma))p_d(x,y) -\sum_{x,y}p_n(y|x)\log(\sigma(-s_{\theta}(x,y)-\gamma))p_d(x)\nonumber\\
= & -\sum_{x,y} \Bigl[\log(\sigma(s_{\theta}(x,y)+\gamma))p_{d}(x,y) +p_n(y|x)\log(\sigma(-s_{\theta}(x,y)-\gamma))p_{d}(x)\Bigr].
\label{eq:app:sub:intro:end}
\end{align}
Next, we consider replacements of $p_{d}(x,y)$ with $p'_{d}(x,y)$ and $p_{d}(x)$ with $p'_{d}(x)$.
By assuming two functions, $A(x,y)$ and $B(x)$, that convert $p_{d}(x,y)$ into $p'_{d}(x,y)$ and $p_{d}(x)$ into $p'_{d}(x)$, we further reformulate Eq.~(\ref{eq:app:sub:intro:end}) as follows:
\begin{align}
& -\sum_{x,y} \Bigl[\log(\sigma(s_{\theta}(x,y)+\gamma))p'_{d}(x,y) +p_n(y|x)\log(\sigma(-s_{\theta}(x,y)-\gamma))p'_{d}(x)\Bigr]\nonumber\\
=&-\sum_{x,y} \Bigl[\log(\sigma(s_{\theta}(x,y)+\gamma))A(x,y)p_{d}(x,y)+p_n(y|x)\log(\sigma(-s_{\theta}(x,y)-\gamma))B(x)p_{d}(x)\Bigr].
\label{eq:app:sub:replaced}
\end{align}
Based on the similar derivation from Eq.~(\ref{eq:ns:loss:kge}) to Eq.~(\ref{eq:app:sub:intro:end}), we can reformulate Eq.~(\ref{eq:app:sub:replaced}) as follows:
\begin{equation}
(\ref{eq:app:sub:replaced}) \approx -\frac{1}{|D|}\sum_{(x,y) \in D} \Bigl[A(x,y)\log(\sigma(s_{\theta}(x,y)+\gamma))+\frac{1}{\nu}\sum_{y_{i}\sim p_n(y_{i}|x)}^{\nu}B(x)\log(\sigma(-s_{\theta}(x,y_i)-\gamma))\Bigr].
\label{eq:app:subsamp}
\end{equation}

\end{document}